\documentclass{article}

\usepackage{arxiv}

\usepackage[utf8]{inputenc} 
\usepackage[T1]{fontenc}    
\usepackage{hyperref}       
\usepackage{url}            
\usepackage{booktabs}       
\usepackage{amsfonts}       
\usepackage{nicefrac}       
\usepackage{microtype}      
\usepackage{lipsum}
\usepackage{graphicx}
\graphicspath{ {./images/} }

\usepackage{graphicx} 
\usepackage{algpseudocode} 
\usepackage{amsmath} 
\usepackage{amssymb} 
\usepackage{algorithm}
\usepackage[numbers]{natbib}
\usepackage{graphicx}
\usepackage{subfigure}
\usepackage{booktabs} 
\usepackage{enumitem} 
\usepackage{tikz} 
\usetikzlibrary{trees, shapes.geometric, positioning} 
\usepackage{indentfirst} 

\title{SMART: A Flexible Approach to Regression using Spline-based Multivariate Adaptive Regression Trees}

\author{
 William Pattie \\
  Northwestern University\\
  \texttt{WilliamPattie2024@u.northwestern.edu} \\
   \And
 Arvind Krishna \\
  Northwestern University\\
  \texttt{krish@northwestern.edu} \\
}
\date{}

\begin{document}
\maketitle
\begin{abstract}

Decision trees are powerful for predictive modeling but often suffer from high variance when modeling continuous relationships. While algorithms like Multivariate Adaptive Regression Splines (MARS) excel at capturing such continuous relationships, they perform poorly when modeling discontinuities. To address the limitations of both approaches, we introduce Spline-based Multivariate Adaptive Regression Trees (SMART), which uses a decision tree to identify subsets of data with distinct continuous relationships and then leverages MARS to fit these relationships independently. Unlike other methods that rely on the tree structure to model interaction and higher-order terms, SMART leverages MARS's native ability to handle these terms, allowing the tree to focus solely on identifying discontinuities in the relationship. We test SMART on various datasets, demonstrating its improvement over state-of-the-art methods in such cases. Additionally, we provide an open-source implementation of our method to be used by practitioners.
\end{abstract}

\section{Introduction}
Decision trees have emerged as a fundamental and widely adopted method in the field of predictive modeling for tabular data. The Classification and Regression Tree (CART) algorithm, introduced by Breiman et al. \cite{breiman2017classification}, has become a seminal method for constructing binary decision trees. CART recursively partitions data into subsets based on feature values, aiming to maximize homogeneity within each node. At each leaf node, when performing regression, it predicts the average response of the training examples that fall into that node. While CART is lauded for its simplicity, interpretability, and ability to capture nonlinear relationships, its accuracy is often limited by its high variance.

\subsection{Background}
To address the high variance problem of decision trees, ensemble methods such as Random Forests \cite{breiman2001random} and Gradient Boosting \cite{friedman2001gbm} have been developed. Random Forests develop many decision trees, each built on a random bootstrapped subset of data and random subset of features. By averaging the predictions of these decorrelated trees, Random Forests tends to reduce the overall variance of the model leading to increased accuracy on unseen data. Gradient Boosting, on the other hand, constructs a model by sequentially adding shallow decision trees, each correcting the errors of its predecessor. Ensemble methods have continued to evolve, with notable advancements such as Bayesian Adaptive Regression Trees (BART) \cite{chipman2010bart}, which builds on boosting by fitting regularized weak learners using a Bayesian backfitting Markov chain Monte Carlo algorithm. Extreme Gradient Boosting (XGBoost) \cite{chen2016xgboost} further enhances boosting with additional regularization and support for weighted and missing data, while CatBoost \cite{prokhorenkova2018catboost} reduces the bias in prediction and in the encoding of categorical variables. While boosting methods are highly flexible, they often create complex weak learners that can result in high variance when modeling intricate continuous functions. In contrast, linear models may offer a more stable and reliable approach for capturing such relationships.

\par Researchers have also explored improving the predictive power of trees by fitting more sophisticated models in the leaf nodes. Early examples such as the M5 algorithm \cite{quinlan1992learning} use the idea of fitting linear models at each leaf node, utilizing all features used in the splits. Similarly, the GUIDE algorithm \cite{loh2002regression} employs a chi-squared analysis of residuals to select splits across individual features and interactions, and fits linear models in the leaves. Zeileis et al. \cite{zeileis2008model} proposed using M-fluctuation tests on generalized linear models to find optimal tree splits, while Rusch et al. \cite{rusch2013gaining} extended this approach by fitting generalized linear models and using M-fluctuations to test for parameter instability and identify optimal partitions. Recent advancements include Local Linear Forests \cite{friedberg2020local}, which use the leaf nodes of decision trees to define adaptive neighborhoods for fitting local linear regressions, thereby improving the smoothness of estimates, and the work of Künzel et al. \cite{kunzel2022linear}, who proposed fitting ridge regressions at leaf nodes of Random Forests. However, these methods still often rely on the tree structure to capture complex interactions between features and higher degree terms, which can lead to suboptimal partitioning.

\subsection{Proposed Method}

\par We introduce Spline-based Multivariate Adaptive Regression Trees (SMART), a novel approach that integrates the strengths of Multivariate Adaptive Regression Splines (MARS) with decision trees. While MARS is a flexible non-parametric regression technique capable of feature selection and modeling continuous non-linear relationships, it is not developed to model discontinuities. In contrast, decision trees are adept at capturing such discontinuities but often exhibit high variance due to their flexibility. SMART addresses these limitations by using decision trees to identify clusters of data with distinct continuous relationships and then applying MARS to fit sophisticated, multivariate spline-based functions within the leaf nodes. This method allows the tree structure to focus solely on identifying distinct continuous relationships, while MARS handles the modeling of interaction and higher-order terms natively. By effectively balancing the modeling of continuous non-linearities, interactions, and discontinuities, SMART presents a unified framework that offers significant improvements over traditional methods that rely on simpler models in the leaf nodes.

\subsection{Contributions and Outline}

\par The main contributions of this paper are:

\begin{itemize}[label=--, leftmargin=18pt, itemsep=2pt, topsep=5pt, partopsep=2pt, parsep=2pt]
    \item A novel method that combines MARS with CART to handle complex datasets with both continuous non-linearities and discontinuities
    \item Experimental results demonstrating the effectiveness of the proposed approach compared to state-of-the-art methods on various datasets
    \item An open-source implementation of SMART for practitioners and the research community to use and build upon \footnote{https://github.com/fyre87/SMART}
\end{itemize}
\par The remainder of this thesis is organized as follows: Section \ref{sec:methodology} presents the methodology, detailing the forward pass, which constructs a continuous function to fit the data, the tree split phase, where the function is recursively partitioned using a decision tree-like approach, and lastly, the pruning process, which simplifies the function at each leaf node. Section \ref{sec:fittingalgorithm} presents an efficient and numerically stable fitting algorithm based on QR decomposition to accelerate the computation of the best split values and variables during the tree split phase. Section \ref{sec:results} discusses the results, comparing SMART to existing approaches such as Random Forest, Gradient Boosting, and MARS. Finally, Sections \ref{sec:conclusion} and \ref{sec:discussion} conclude the thesis, summarizing the key findings and outlining potential future research directions to further enhance the proposed method.

\section{Methodology}
\label{sec:methodology}

Let $X \in \mathbb{R}^{n \times m}$ be the matrix that holds the training features, $X_{i,*}$ be the $i$th observation of the data, and $X_{*,j}$ be the $j$th feature column of the data. Let $(X_{i,*}, Y_i)_{i=1}^{n}$ represent the $n$ observations in the training data where $Y \in \mathbb{R}^n$ is the dependent variable vector.

Let us consider an example that has both discontinuities and continuous non-linearities in the relationship. Generate $X_{i,1} \in \mathcal{U}(0, 6)$ for $i \in \{1 \dots 200\}$ and consider the following function:
\begin{equation}
\label{visual_equation}
Y_i = \epsilon_i + 
\begin{cases}
\text{sin}(\pi X_{i,1}) \text{ if } X_{i,1}\in[0,2)\\
4X_{i,1} \text{ if } X_{i,1}\in[2,4) \\
0.2\text{exp}(X_{i,1}-3) \text{ otherwise}
\end{cases}
\end{equation}
where $\epsilon_i \sim \mathcal{N}(0, 1)$ is the random error term.

\begin{figure}[h]
\centering
\includegraphics[width=0.7\textwidth]{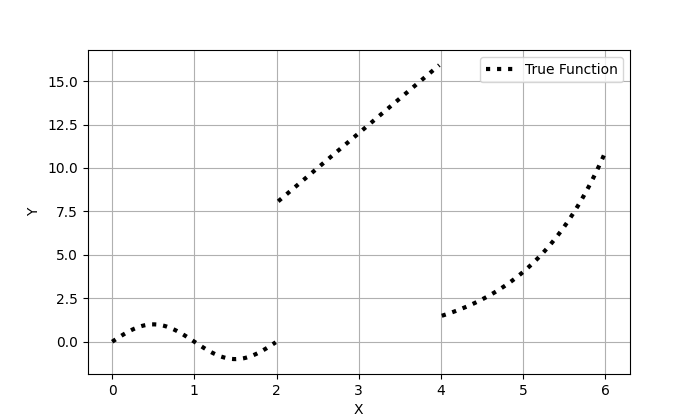}
\caption{Equation \ref{visual_equation} mapped. The noise is not visualized.}
\label{fig:blank}
\end{figure}

\begin{figure}[ht]
  \centering
  \includegraphics[width=0.49\textwidth]{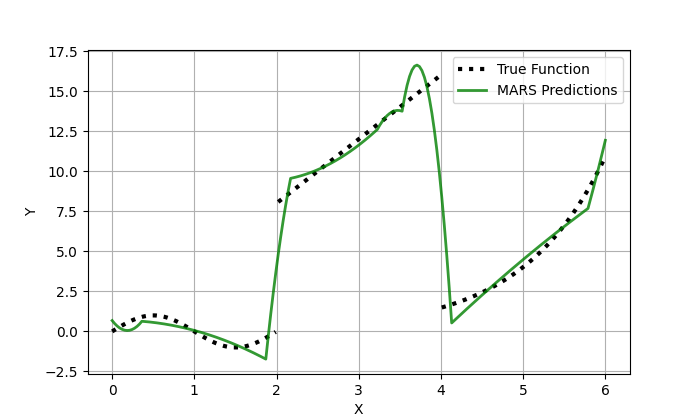}
  \includegraphics[width=0.49\textwidth]{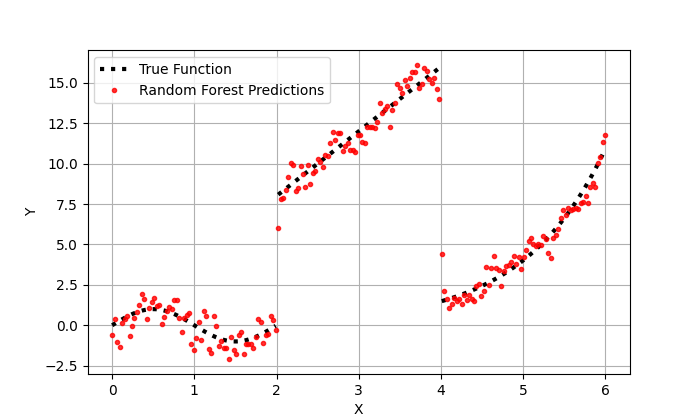}
  \caption{On the left is MARS fit with a max degree of interaction of 2. On the right Random Forest is tuned using grid search and 5-fold cross-validation. MARS has an RMSE of 1.11 on the true function. Random Forest has an RMSE of 0.71 on the true function.}
  \label{fig:MARS and RF}
\end{figure}
This function is visualized in figure \ref{fig:blank}. In figure \ref{fig:MARS and RF} we fit both MARS and Random Forests on the data. The MARS algorithm had a high bias due to its inability to model discontinuities, while Random Forests had high variance due to their flexible structure. Our algorithm attempts to balance the low bias of Random Forests and the low variance of MARS to achieve better accuracy. In the remainder of the section, we will describe in detail the three phases of our algorithm and show how the fit on Equation \ref{visual_equation} evolves. 

\subsection{Forward Pass}

The forward pass of the algorithm starts with the MARS forward pass algorithm. Start with just an intercept term $f(X) = B_0$. Then consider adding ‘reflected pairs’ of piecewise linear basis functions of the form $\beta_i(X_{*,j} - t)_+ + \beta_{i+1}(t - X_{*,j})_+$ for parameters $\beta_i$, $\beta_{i+1}$, and some constant $t$ where:
\[
(x - t)_+ = \left\{
\begin{array}{ll}
    x-t & \text{if } (x-t) > 0, \\
    0 & \text{otherwise.}
\end{array}
\right.
\]
and $(X_{*,j} - t)_+$ is the same function applied to the whole column:
\[
(X_{*,j} - t)_+ = [(X_{1,j}-t)_+,\dots,(X_{n,j}-t)_+]^T
\]
The algorithm finds $t$ and $j$ which form the reflected pair which minimizes the residual sum of squares (RSS) as defined by:
\[
RSS = \sum_{i=1}^{n}(Y_i-f(X_{i,*}))^2 
\]
If, however, the first order term $\beta_iX_{*,j}$ results in a lower generalized cross-validated error (GCV) than the reflected hinge functions $\beta_i(X_{*,j} - t)_+ + \beta_{i+1}(t - X_{*,j})_+$, the first order term is added instead of the basis function pair. The GCV, which is similar to RSS but penalizes for adding more parameters, is defined by:
\begin{equation}
\label{GCV}
GCV(\lambda) = \frac{\sum_{i=1}^{n}(Y_i-f(X_{i,*}))^2}{(1-M(\lambda)/n)^2}
\end{equation}
where $M(\lambda)$ is the effective number of parameters in the model as described in Hastie et al \cite{hastie2009elements}.
\par After the first term is added, the algorithm continues to consider adding reflected pairs of hinge functions or first order terms. Additionally, it considers interaction terms, which are products of an existing term in the model with a new hinge function or first order term. The algorithm chooses whichever option lowers the RSS the most, and this process of adding terms and interactions continues until the change in the RSS is too small or until the maximum number of terms hyperparameter is reached. The result is a function of the form:
\begin{equation}
\label{MARS EQ}
f(X) = \beta_0 + \sum_{m=1}^{M}\beta_mh_m(X)
\end{equation}
where each $h_i(X)$ is of one of the following forms: $X_{*,j}$, $(X_{*,j} - t)_+$, $(t- X_{*,j})_+$, or a product of two or more of the previous options.

\begin{figure}[h]
\centering
\includegraphics[width=0.7\textwidth]{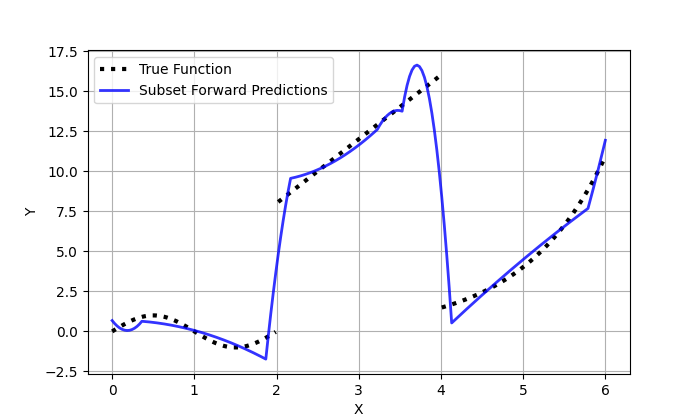}
\caption{The result of the forward pass on Equation \ref{visual_equation}. The model was fit with a max degree of interaction of 2. The model currently has a RMSE of 1.12 on the true function. }
\label{fig:forward}
\end{figure}
Figure \ref{fig:forward} shows the model fit on Equation \ref{visual_equation} after the forward pass. The fit suffers from the same problem as MARS in Figure \ref{fig:MARS and RF}, where the inductive bias prevents it from fitting the jump discontinuities. To improve on this, we allow the model to consider adding decision tree splits as described in the following phase.

\subsection{Tree Split Algorithm}
\par The goal of the splitting algorithm is to find a variable $d \in \{1, \dots m\}$ and a split value $s$, such that the partition
\begin{equation}
\label{Partition EQ}
L_{d,s} := \{(X_{i,*}\text{, }Y_i) \mid X_{i,d} \leq s\}\text{, } R_{d,s} := \{(X_{i,*}\text{, }Y_i) \mid X_{i,d} > s\}
\end{equation}
results in the maximum reduction in the RSS on unseen data once Equation \ref{MARS EQ}’s coefficients are recomputed on each of the two partitions. 
\par To achieve this, we start by randomly splitting the training data into $70\%$ fitting data and $30\%$ validation data. Then, to find the optimal splitting value and variable, loop through every unique value within every variable in the fitting data, split both the fitting dataset and the validation dataset according to Equation \ref{Partition EQ}, fit Equation \ref{MARS EQ} on the fitting data, and then calculate the RSS on the validation data. The use of the validation set helps ensure the choice of split creates a model that works well on unseen data and doesn't overfit the training data. We then record the split variable and split value which resulted in the lowest validation loss in terms of RSS. We formalize this process in Algorithm \ref{Best Split Finder}. 

\begin{algorithm}[H]
\caption{Best Split Finder for Continuous Variables}
\label{Best Split Finder}
\textbf{Input: } Features $X \in \mathbb{R}^{n\times m}$. Dependent variable vector $Y \in \mathbb{R}^{n}$ \newline
\textbf{Output: } Best split variable and value
\begin{algorithmic}[1]

\State $Best\_RSS \gets \infty$
\State Initialize $f_L$ and $f_R$ using Equation \ref{MARS EQ}
\State Split 70\% of $(X_{i,*}, Y_i)_{i=1}^{n}$ into $(X^F_{i,*}, Y^F_i)^{n_F}_{i=1}$ and 30\% into $(X^V_{i,*}, Y^V_i)^{n_V}_{i=1}$
\For{$d \in \{1, \ldots, m\}$}
    \For{$s \in X^F_{*,d}$}
        \State Partition the dataset
        \State $L^F_{d,s}:= \{(X^F_{i,*}\text{, }Y^F_i) \mid X^F_{i,d} \leq s\}$, $R^F_{d,s}:= \{(X^F_{i,*}\text{, }Y^F_i) \mid X^F_{i,d} > s\}$
        \State $L^V_{d,s}:= \{(X^V_{i,*}\text{, }Y^V_i) \mid X^V_{i,d} \leq s\}$, $R^V_{d,s}:= \{(X^V_{i,*}\text{, }Y^V_i) \mid X^V_{i,d} > s\}$
        \State Update $f_L$ using $L^F_{d,s}$, and $f_R$ using $R^F_{d,s}$
        \State 
        \State $RSS = RSS_L + RSS_R$
        \If{$RSS < Best\_RSS$}
            \State $Best\_RSS \gets RSS$
            \State $Best\_VARIABLE \gets d$
            \State $Best\_SPLIT \gets s$
        \EndIf
    \EndFor
\EndFor
\end{algorithmic}
\end{algorithm}

\par The repeated splitting process can be done naively by running ordinary least squares (OLS) regression of Equation \ref{MARS EQ} on the fitting data. However, we use a faster algorithm to compute a least squares solution as described in Section \ref{sec:fittingalgorithm}. 
\par After running the algorithm, we have a potential split value and variable which can be used to create a split of the form of Equation \ref{Partition EQ}. Rather than proceeding with the split immediately, however, we first calculate the 5-fold cross-validation RSS of the original model, and the 5-fold cross-validation RSS of the considered partition. Using cross-validation at this stage provides a more robust estimate of the split's impact on the model's ability to generalize to unseen data, reducing the risk of overfitting to a specific validation set. Moreover, by setting a pre-defined threshold for improvement in cross-validation RSS (in our case, 1\%), we introduce a regularization mechanism that prevents the tree from growing too deep and overfitting the data. If the cross-validated error improvement meets or exceeds the threshold, the split is confirmed and the splitting process repeats itself recursively on the resulting child nodes. Otherwise, the split is not confirmed, and the current node becomes a terminal node. \footnote{If there are more terms in Equation \ref{MARS EQ} than observations left in the training data in the current node, we make the node a terminal node before running Algorithm \ref{Best Split Finder} as there may not be enough data to run cross-validation. Also, if there are few points in a node (defined by $n<10(M+1)$ where $M$ is the number of added parameters from equation \ref{MARS EQ}), there's not enough data to create a validation set of sufficient size, so we fit and validate on the training data instead to find the best variable and split.} 
\par The result is a function of the form
\begin{equation}
\label{final equation}
f(X) = f_1(X) + ... + f_k(X)
\end{equation}
where each $f_i(X)$ represents a terminal node equation of the form of Equation \ref{MARS EQ} for all datapoints in its node, and 0 for all datapoints outside its node, and $k$ represents the total number of terminal nodes. 

\begin{figure}[h]
\centering
\includegraphics[width=0.7\textwidth]{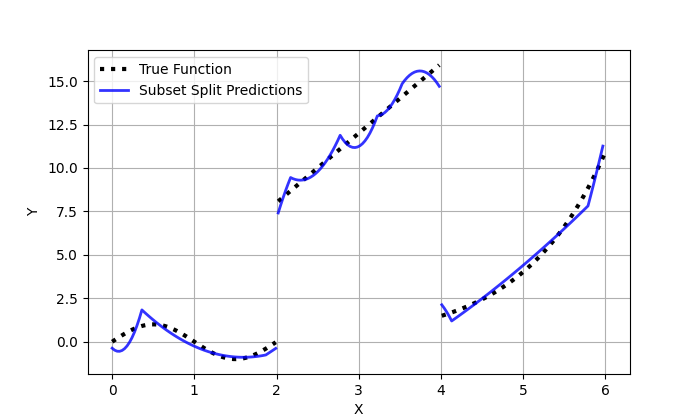}
\caption{The result of the tree split phase on Equation \ref{visual_equation}. The model's RMSE reduced from 1.12 to 0.42 on the true function.}
\label{fig:split}
\end{figure}

Figure \ref{fig:split} shows what the fit on Equation \ref{visual_equation} looks like after the tree splits are made. In this case, the model made splits at the jump discontinuities and significantly improved the fit on the true data.

\subsection{Pruning Pass}
\par In the MARS algorithm, the forward pass typically creates an overfit model \cite{hastie2009elements} and so a backward pruning pass is used. The pruning stage of MARS prunes the terms of the model one by one, deleting the term that results in the lowest mean squared error post-deletion until the entire model is pruned. Upon completion, the algorithm returns the equation that resulted in the lowest GCV (as defined in Equation \ref{GCV}), and the MARS algorithm concludes. 
\par Each of the partitions created in Algorithm \ref{Best Split Finder} contains potentially its own unique distribution. We use the traditional MARS pruning algorithm on each of the leaf nodes individually, allowing the functions in each leaf node to have their own terms and structure. This process is formalized in Algorithm \ref{Pruning Algorithm}. 

\begin{algorithm}[H]
\caption{Pruning Algorithm}
\label{Pruning Algorithm}

\textbf{Input: } $f(X) = f_1(X) + ... + f_k(X)$ as in Equation (\ref{final equation}) \newline
\textbf{Output: }  $f(X) = f_1(X) + ... + f_k(X)$ with each leaf node pruned
\begin{algorithmic}[1]
\For{$i \in \{1 \dots k\}$}
    \State Prune $f_i$ using the traditional MARS pruning algorithm
\EndFor
\end{algorithmic}
\end{algorithm}

\begin{figure}[h]
\centering

\includegraphics[width=0.70\textwidth]{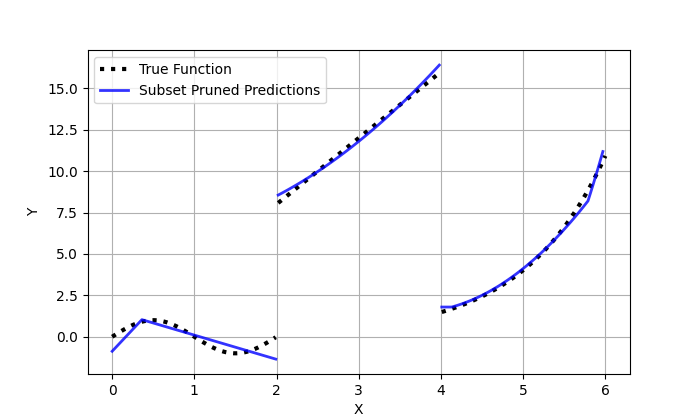}

\caption{The result of the tree split phase on Equation \ref{visual_equation}. The model's RMSE reduced from 0.42 to 0.32 on the true function.}
\label{fig:pruned}
\end{figure}

Figure \ref{fig:pruned} shows what the fit on Equation \ref{visual_equation} looks like after the pruning algorithm is applied. In this case, the pruning improved the error on the true function. The final RMSE on the true function, 0.32, is much lower than that of the Random Forest, 0.71, and that of the MARS, 1.11. 

\section{Fitting Algorithm}
\label{sec:fittingalgorithm}
\par To find the optimal splitting value and variable in Algorithm \ref{Best Split Finder}, it is possible to loop through every possible variable and every unique value within that variable and compute the RSS of the resulting fit of Equation \ref{MARS EQ} on the validation data using OLS. However, if $X_{*,d}$ is a continuous variable, then there are up to $n_F-1$ unique split values to evaluate where $n_F$ is the number of datapoints in the fitting data. In other words, sort the potential split value of $d$, or the values of $X_{*,d}$ in ascending order: $\{s_1,\dots, s_{n_F}\}$ such that $s_i \leq s_j$ for $i<j$. If $d$ is continuous, then the partition:
\[
 L_{d,s_j} = \{(X^F_{i,*}\text{, }Y^F_i) \mid X^F_{i,d} \leq s_j\}
\]
and the partition:
\[
 L_{d,s_{j+1}} = \{(X^F_{i,*}\text{, }Y^F_i) \mid X^F_{i,d} \leq s_{j+1}\}
\]
could differ by as few as 1 datapoint. Given this, it would be inefficient to solve the ordinary least squares equation on $L_{d,s_j}$, and then entirely recompute the coefficients on $L_{d,s_{i+j}}$.
\par Several online learning methods exist for incrementally computing the least squares solutions. The recursive least squares (RLS) adaptive filter algorithm provides an efficient updating algorithm for computing the least squares coefficients \cite{haykin2002adaptive}. However, in practice, numerical stability issues arise with RLS as roundoff errors accumulate in areas of finite precision \cite{liavas1999numerical}. Another method that avoids stability issues is stochastic gradient descent (SGD) which updates the model parameters iteratively by taking small steps in the direction of the negative gradient of the loss function. However, SGD can be sensitive to the choice of learning rate and may require careful tuning to achieve optimal performance \cite{bengio2012practical}, which can significantly increase tuning and training time. 
\par Considering these limitations, we use a QR decomposition updating method. QR decomposition can solve a least squares problem:
 \[
  A\vec{x} = \vec{b}
 \]
for the coefficients $\vec{x}$ by decomposing $A$ into the product of an orthogonal square matrix $Q$ and an upper triangular (or upper trapezoidal for under-determined systems) matrix $R$, and then recalling that for orthogonal matrices, $Q^{-1} = Q^T$. One can then rewrite the least squares problem as:
\[
  R\vec{x} = Q^T\vec{b}
\]
then solve for the coefficients $\vec{x}$ by solving $R^{-1}Q^T\vec{b}$ or by using back substitution. 
\par We use Algorithm 6 in Hammarling et al. \cite{hammarling2008updating} which allows for space-efficient, time-efficient, and stable recomputation of the $R$ and $Q^T\vec{b}$. The algorithm uses Givens rotations to update the matrices. Let $W$ represent the current solution for $Q^T\vec{b}$. To introduce the notation, let $X_{k, i:i+2}$, for instance, refer to a subset of the matrix $X$ including the kth row and the i, i+1, and i+2 columns, and let $M$ refer to the number of variables created in the forward pass as in equation \ref{MARS EQ}. Then the update method is formalized in Algorithm \ref{Updating QR Algorithm}. 

\begin{algorithm}[H]
\label{Updating QR Algorithm}
\caption{Updating QR using Givens rotations}

\textbf{Input: } $m_{*} = \textbf{min}(n, M)$ \newline
$W\gets \text{current } Q^T\vec{b}\in \mathbb{R}^{m_* \times 1}$ \newline
$R\gets \text{current upper triangular } R\in \mathbb{R}^{m_* \times M}$\newline
$X_{n+1, *}\in \mathbb{R}^{1 \times M}$ as new X row \newline
$Y_{n+1}\in \mathbb{R}$ as new response\newline
\textbf{Output: } Updated $W$ and $R$ matrices

\begin{algorithmic}[1]


\For{$j \in \{1 \dots m_{*}\}$}
    \State c, s = \textbf{Givens}($R_{j, j}$, $X_{n+1, j}$)
    \State $R_{j, j} = c(R_{j, j}) - s(X_{n+1, j})$ 
    \State Update jth row of $R$ and jth column of $X_{n+1, *}$:
        \State \hspace{0.8em} $t1 =  R_{j, j+1:m_{*}}\in \mathbb{R}^{1 \times (m_{*}-j)}$
        \State \hspace{0.8em} $t2 = X_{n+1, j+1:m_{*}}\in \mathbb{R}^{1 \times (m_{*}-j)}$
        \State \hspace{0.8em} $R_{j, j+1:m_{*}} = c(t1) - s(t2)$
        \State \hspace{0.8em} $X_{n+1, j+1:m_{*}} = s(t1) + c(t2)$
    \State Update $Y_{n+1}$ jth row of d:
        \State \hspace{0.8em} $t1 = w_{j, 1}$
        \State \hspace{0.8em} $t2 = y_{n+1}$
        \State \hspace{0.8em} $W_{j, 1} = c(t1) - s(t2)$
        \State \hspace{0.8em} $Y_{n+1} = s(t1) + c(t2)$
\EndFor
 
\If {$n < M$}
    \State If less rows in R than columns, add a row to R. 
    \State $R\gets \begin{bmatrix} R \\ X_{n+1, *} \end{bmatrix} \in \mathbb{R}^{n+1 \times M}$
    \State $W\gets \begin{bmatrix} W \\ Y_{n+1} \end{bmatrix} \in \mathbb{R}^{n+1 \times 1}$
\EndIf

\end{algorithmic}
\end{algorithm}

where the Givens function is given by Algorithm 1.1 in Hammarling et al. \cite{hammarling2008updating}:

\begin{algorithm}[H]
\caption{Givens}
\label{Givens Algorithm}
\textbf{Input: } Scalar inputs $a$ and $b$ \newline
\textbf{Output: } Scalar outputs $c$ and $s$
\begin{algorithmic}[1]

    \If {$b = 0$}
    \State $c = 1, s = 0$
    \ElsIf{$|b| \geq |a|$}
    \State $t = -a/b\text{, }s = 1/\sqrt{1+t^2}\text{, }c = st$
    \Else
    \State $t = -b/a\text{, }c = 1/\sqrt{1+t^2}\text{, }s = st$
    \EndIf

\end{algorithmic}
\end{algorithm}

\par Finally, if $d$ is a binary or categorical variable, it is computationally more efficient and numerically stable to compute the RSS using OLS at each of the few split values, not to have to update QR and recompute the RSS for every observation. Thus, in the case of a categorical variable, we opt for OLS at each split value to find the best split value. 

\subsection{Time Complexity}

\par Algorithm \ref{Givens Algorithm} consists of nothing but scalar multiplication, and so can be completed in O(1) time. 
\par In Algorithm \ref{Updating QR Algorithm}, lines 7-8 and 14-18 consist of scalar multiplication which can all be completed in O(1) time. Lines 10-13 multiply scalars ($c$ and $s$) by vectors ($t1$ and $t2$) of size $m_*-j$ which involves up to $m_*$ scalar multiplications and can be completed in O($m_*$) time. Thus the whole block from lines 7-18 has O($m_*$) time complexity, and this block is repeated $m_*$ times meaning that lines 6-19 have O($m_*^2$)=O($\textbf{min}(n,M)^2$)=O($M^2$) time complexity. 
\par In line 22-23, given the maximum final size of the $R$ and $W$ matrices is ($R\in\mathbb{R}^{M\times M}$, $W\in\mathbb{R}^{M\times 1}$), the new matrices require, in the worst case scenario, allocating a $M$ by $M$ array into memory which takes O($M^2$) time. Thus all of algorithm \ref{Updating QR Algorithm} can be completed in O($M^2+M^2$)=O($M^2$) time. 
\par After each run of algorithm \ref{Updating QR Algorithm}, we must make a prediction to compute the validation loss given our updated $R$ and $W$ matrices using back substitution. Given that $R$ is of size $\mathbb{R}^{M\times M}$ in the worst case, the back substitution algorithm runs in O($M^2$) \cite{cormen2022introduction}. Thus together, the time required to run algorithm \ref{Updating QR Algorithm} and run the resulting back substitution is O($M^2 + M^2$) = O($M^2$). 
\par For a single continuous variable, the tree split algorithm requires running algorithm \ref{Updating QR Algorithm} and back substitution once per observation in the dataset ($n$) in the worst-case scenario. Thus the resulting time complexity to find the best tree split along a continuous variable is O($nM^2$). 

\section{Results}
\label{sec:results}
\subsection{Friedman tests}
\par In this section, we evaluate the performance of SMART by comparing it against several well-established methods, including Random Forests, Multivariate Adaptive Regression Splines (MARS), XGBoost, and Local Linear Forests. The experiments are conducted on various simulated datasets to assess the model's effectiveness under different scenarios. We use the Scikit-learn Python package \cite{kramer2016scikit} to implement Random Forests and the dlmc XGBoost package \cite{chen2016xgboost} to implement XGBoost, also in Python. We use the R package grf \cite{grf2019} to implement Local Linear Forests. To ensure a fair comparison, we use 5-fold cross-validation to tune the hyperparameters in a window around their default values (details can be found in Appendix A). 
\par To fit Multivariate Adaptive Regression Splines, we use the py-earth Python package \cite{py-earth}. For the following Friedman datasets, we use a max degree of interaction of 2 for MARS and SMART, a max parameters hyperparameter $M+1$ of 100, and we do not tune the hyperparameters of either model.  

\par For our first experiment, we use the Friedman 1 dataset \cite{friedman1991mars} which is a commonly used dataset for evaluating non-parametric regression techniques \cite{friedberg2020local}, \cite{matthew2015bayesian}. We generate $X_{*,1}\dots X_{*,d}$ independent and identically distributed variables from the uniform distribution $\mathcal{U}(0, 1)$. Then, the response $Y$ is determined by: 
\[
\label{Friedman1 EQ}
Y_i = 10\text{sin}(\pi X_{i,1} X_{i,2}) + 20(X_{i,3} - 0.5)^2 + 10X_{i,4} + 5X_{i,5} + \epsilon_i
\]
where $\epsilon_i \sim \mathcal{N}(0, \sigma^2)$. This relationship tests how well a model can fit interaction terms ($10\text{sin}(\pi X_{i,1} X_{i,2})$), quadratic terms ($20(X_{i,3} - 0.5)^2$), linear terms ($10X_{i,4} + 5X_{i,5}$), and how well it can ignore the added noise variables ($X_{*,6}\dots X_{*,d}$). 
\par The setup we chose is similar to the setup used by Friedberg et al \cite{friedberg2020local}, which chooses $\sigma$ of 5 or 20, a $d$ of 10, 30, or 50, and training datasets of size 1000 or 5000 simulated points. To evaluate the test RMSE, we used the same input values as the training data and generated the corresponding true function values (with $\epsilon = 0$).

\begin{table}[t]
\centering

\label{tab:friedman1}
\begin{tabular}{@{}ccc|ccccc@{}}
\toprule
\( d \) & \( n \) & \( \sigma \) & RF & LLF & XGBoost & MARS & SMART \\
\midrule
10 & 1000 & 5 & 3.39 & 2.57 & 1.86 & \textbf{1.74} & \textbf{1.74} \\
10 & 5000 & 5 & 2.46 & 2.07 & 1.19 & \textbf{0.69} & \textbf{0.69} \\
30 & 1000 & 5 & 3.32 & 3.10 & \textbf{2.27} & 2.60 & 2.60 \\
30 & 5000 & 5 & 3.22 & 3.01 & 1.39 & \textbf{0.87} & \textbf{0.87} \\
50 & 1000 & 5 & 3.29 & 3.45 & \textbf{2.28} & 2.72 & 3.21 \\
50 & 5000 & 5 & 3.25 & 3.29 & 1.45 & \textbf{1.10} & \textbf{1.10} \\
\midrule
10 & 1000 & 20 & 10.87 & 8.19 & \textbf{6.32} & 6.70 & 6.70 \\
10 & 5000 & 20 & 6.60 & 3.10 & 3.16 & \textbf{3.02} & \textbf{3.02} \\
30 & 1000 & 20 & 12.87 & 8.20 & \textbf{7.41} & 9.91 & 9.91 \\
30 & 5000 & 20 & 8.74 & 3.15 & \textbf{3.10} & 4.49 & 4.49 \\
50 & 1000 & 20 & 12.28 & 12.03 & \textbf{7.38} & 11.43 & 10.00 \\
50 & 5000 & 20 & 8.56 & 11.08 & \textbf{3.63} & 5.56 & 5.56 \\
\midrule
& & Average & 6.57 & 5.27 & \textbf{3.45} & 4.24 & 4.16 \\
\bottomrule
\end{tabular}
\caption{RMSE on Friedman 1.}
\end{table}

\par We measure the RMSE on the test set in Table 1. From the table, the test error from SMART and MARS was identical for most cases. This is to be expected as the MARS forward pass is already designed to include interaction terms, quadratic terms, linear terms, and feature selection. SMART successfully avoided making unnecessary tree splits here and captured the benefits of the MARS algorithm. 
\par We repeat a similar testing scheme for the other Friedman 2 and Friedman 3 simulations. Let $X_{i,1} \sim \mathcal{U}(0, 100)$, $X_{i,2} \sim \mathcal{U}(40\pi, 560\pi)$, $X_{i,3} \sim \mathcal{U}(0, 1)$, $X_{i,4} \sim \mathcal{U}(1, 11)$, and $\epsilon_i \sim \mathcal{N}(0, \sigma^2)$. Then the response for Friedman 2 is defined by:
\[
Y_i = \sqrt{X_{i,1}^2 + \left(X_{i,2}X_{i,3} - \frac{1}{X_{i,2}X_{i,4}}\right)^2} + \epsilon_i
\]
and the response for Friedman 3 is defined by:
\[
Y_i = \arctan\left(\frac{X_{i,2}X_{i,3} - \frac{1}{X_{i,2}X_{i,4}}}{X_{i,1}}\right) + \epsilon_i
\]
\par Similar to Friedman 1, we simulate the models with $\sigma$ of 5 or 20, training dataset sizes of 1000 or 5000, and generate the same points with no noise for the test data. 

\begin{table}[t]
\centering

\label{tab:friedman2}
\begin{tabular}{@{}cc|ccccc@{}}
\toprule
\( n \) & \( \sigma \) & RF & LLF & XGBoost & MARS & SMART \\
\midrule
1000 & 5 & 7.29 & \textbf{2.92} & 4.27 & 9.67 & 3.17 \\
5000 & 5 & 4.22 & \textbf{1.94} & 3.61 & 9.92 & 2.30 \\
1000 & 20 & 14.59 & \textbf{5.43} & 14.97 & 6.09 & 6.01 \\
5000 & 20 & 12.54 & 4.53 & 8.50 & 5.99 & \textbf{3.69} \\
\midrule
& Average & 9.66 & 3.71 & 7.84 & 7.92 & \textbf{3.69} \\
\bottomrule
\end{tabular}
\caption{RMSE on Friedman 2.}
\end{table}

\begin{table}[t]
\centering

\label{tab:friedman3}
\begin{tabular}{@{}cc|ccccc@{}}
\toprule
\( n \) & \( \sigma \) & RF & LLF & XGBoost & MARS & SMART \\
\midrule
1000 & 5 & 1.86 & \textbf{0.50} & 0.72 & 0.60 & 0.60 \\
5000 & 5 & 0.91 & 0.46 & 0.43 & \textbf{0.24} & \textbf{0.24} \\
1000 & 20 & 7.45 & \textbf{1.67} & 2.12 & 2.05 & 2.05 \\
5000 & 20 & 3.52 & 1.98 & 0.98 & \textbf{0.34} & \textbf{0.34} \\
\midrule
& Average & 3.44 & 1.15 & 1.06 & \textbf{0.81} & \textbf{0.81} \\

\bottomrule
\end{tabular}
\caption{RMSE on Friedman 3.}
\end{table}

\par The Friedman 3 results in Table 3 show similar results to Friedman 1. SMART did not find any splits that improved the cross-validation error by enough to execute but it was able to capture the high performance of the MARS model and tie for the best performance on average. On Friedman 2 in Table 2, however, the model found splits to improve the fit over the traditional MARS algorithm dramatically and thus had the lowest average RMSE of all models. 

\subsection{Piecewise Test}
\label{subsec:PiecewiseTest}
\par The next experiment aimed to evaluate SMART's ability to capture piecewise components in a continuous function, a key feature of the tree split phase in the proposed algorithm. For this test, we generated 5 synthetic datasets with three to five terms of the form $a_iX_{*,i}^n$ for $n\in \{1,2,3\}$, one to two hinge terms of the form $a_i(X_{*,i}-b_i)_+ - a^*_i(b_i - X_{*,i})_+$, one logarithmic term of the form $a_i\text{log}(X_{*,i} + 1)$, and one interaction term. All of the coefficients $a_i$ are randomly chosen from the uniform distribution $\mathcal{U}(-4, -0.4) \cup \mathcal{U}(0.4, 4)$. To introduce a piecewise component, we randomly selected a variable and a corresponding threshold value that partitions the dataset into two subsets: one containing 90\% of the data and the other with the remaining 10\%. For the smaller subset, we then re-randomized all the coefficients associated with the data points, effectively creating a distinct function for that partition. We generated 5 synthetic functions each with a noise of $\mathcal{N}(0, 10^2)$, 5,000 training data points, and a test set of the same size and points with no noise. More details on the specific generation process can be found in Appendix B. 

\par Given the functions are cubic, we chose to fit the MARS and SMART with a maximum degree of interaction hyperparameter of 3, and we kept the same hyperparameter tuning specifications as before for the other models. 

\begin{table}[t]
\centering

\label{tab:synthetic}
\begin{tabular}{@{}c|ccccc@{}}
\toprule
 & RF & LLF & XGBoost & MARS & SMART \\
\midrule
Synthetic 1 & 5.47 & 5.03 & 2.72 & 2.54 & \textbf{1.56} \\
Synthetic 2 & 5.80 & 4.46 & 3.94 & 3.78 & \textbf{1.51} \\
Synthetic 3 & 6.56 & 5.24 & 2.89 & 2.50 & \textbf{2.38} \\
Synthetic 4 & 6.67 & 5.42 & 2.77 & 2.37 & \textbf{1.66} \\
Synthetic 5 & 6.36 & 5.89 & 3.06 & 4.67 & \textbf{1.43} \\
\midrule
Average & 6.17 & 5.21 & 3.08 & 3.17 & \textbf{1.71} \\

\bottomrule
\end{tabular}
\caption{RMSE on Synthetic.}
\end{table}

\begin{table}[t]
\centering

\label{tab:synthetic_splits}
\begin{tabular}{@{}c|ccccc@{}}
\toprule
& True Var & True Split & SMART Var & SMART Split \\
\midrule
Synthetic 1 & 1 & 1.60 & 1 & 1.60 \\
Synthetic 2 & 2 & 8.00 & 2 & 8.00 \\
Synthetic 3 & 3 & 8.00 & 3 & 7.24 \\
Synthetic 4 & 1 & -2.40 & 1 & -2.40 \\
Synthetic 5 & 5 & 8.00 & 5 & 8.00 \\

\bottomrule
\end{tabular}

\begin{minipage}{\textwidth}
\footnotesize
\begin{center}
\caption{True Split vs SMART Split.}
\textit{Note:} In all cases, SMART only made one split as desired.
\end{center}
\end{minipage}
\end{table}

\par From Table 4, SMART was the most accurate of all models in all cases. The performance came from the model's ability to identify the correct split and adjust its coefficients accordingly. This is demonstrated in Table 5, where our model's chosen split aligns at or almost at the true functions split for all cases.

\subsection{Pruning Test}
\label{subsec:PruningTest}

For the final simulation, we test SMART's ability to find successive splits and to prune each leaf node to match the leaf node's structure. 

\begin{figure}[htb]
    \centering
\begin{tikzpicture}[
    decision/.style={diamond, draw, text width=4em, text badly centered, inner sep=1pt},
    leaf/.style={rectangle, draw, text centered, inner sep=5pt},
    edge from parent/.style={draw, edge from parent path={(\tikzparentnode) -- (\tikzchildnode)}},
    level 1/.style={sibling distance=8cm},
    level 2/.style={sibling distance=4cm},
    level distance=3cm
]
\node [decision] {$X_{i,4} > 0$}
    child { node [decision, below left=2cm and 2cm of \tikzparentnode] {$X_{i,2} > 0$}
        child { node [leaf, below left=1cm and 0cm of \tikzchildnode] {$Y_i = X_{i,1} + \epsilon_i$} edge from parent node [left] {yes} }
        child { node [leaf, below right=1cm and -0.5cm of \tikzchildnode] {$Y_i = X_{i,2} + \epsilon_i$} edge from parent node [right] {no} }
        edge from parent node [left] {yes}
    }
    child { node [decision, below right=2cm and 2cm of \tikzparentnode] {$X_{i,1} > 0$}
        child { node [leaf, below left=1cm and -0.5cm of \tikzchildnode] {$Y_i = X_{i,3} + \epsilon_i$} edge from parent node [left] {yes} }
        child { node [leaf, below right=1cm and 0cm of \tikzchildnode] {$Y_i = X_{i,4} + \epsilon_i$} edge from parent node [right] {no} }
        edge from parent node [right] {no}
    };
\end{tikzpicture}
    \caption{Decision tree for determining $Y_i$.}
    \label{fig:decision_tree}
\end{figure}
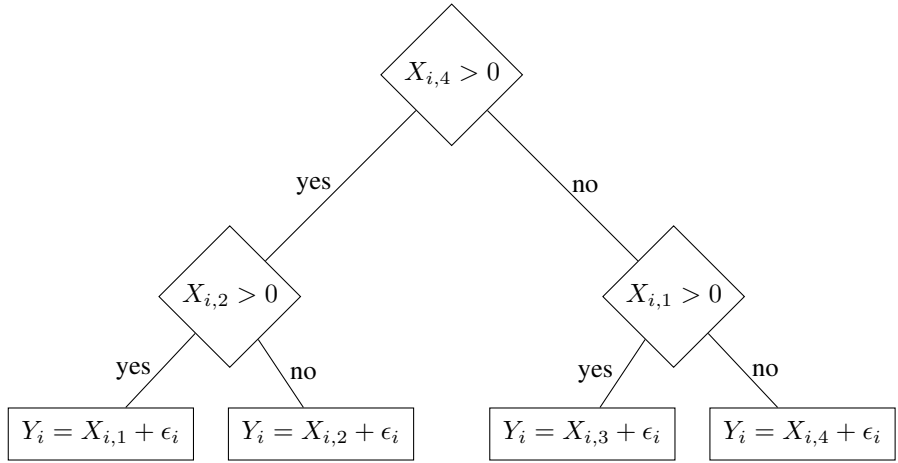

\par Generate $X_{*,1}\dots X_{*,4}$ independent and identically distributed variables from the uniform distribution $\mathcal{U}(-2, 2)$. We simulate 20,000 datapoints where figure \ref{fig:decision_tree} shows how the response variable is created. The structure has a simple linear regression at each of the terminal nodes with added noise. The noise terms $\epsilon_i$ are independent and identically distributed according to the standard normal distribution $\mathcal{N}(0, 1)$. 

\begin{figure}[htb]
    \centering
    \begin{tikzpicture}[
        decision/.style={diamond, draw, text width=4em, text badly centered, inner sep=1pt},
        leaf/.style={rectangle, draw, text centered, inner sep=5pt, text width=9em, minimum height=1.5em},
        edge from parent/.style={draw, edge from parent path={(\tikzparentnode) -- (\tikzchildnode)}},
        level 1/.style={sibling distance=8cm},
        level 2/.style={sibling distance=4cm},
        level distance=3cm
    ]
\node [decision] {$X_{i,4} > 0$}
    child { node [decision, below left=2cm and 2cm of \tikzparentnode] {$X_{i,2} > 0.01$}
        child { node [leaf, below left=1cm and 0cm of \tikzchildnode] {$Y_i = -0.97 + 0.99(X_{i,1}+0.96)_+ - 0.97(-0.96-X_{i,1})_+$} edge from parent node [left] {yes} }
        child { node [leaf, below right=2cm and -1cm of \tikzchildnode] {$Y_i = 0.96X_{i,2} - 0.02$} edge from parent node [right] {no} }
        edge from parent node [left] {yes}
    }
    child { node [decision, below right=2cm and 2cm of \tikzparentnode] {$X_{i,1} > 0.01$}
        child { node [leaf, below left=1cm and -1cm of \tikzchildnode] {$Y_i = 0.97X_{i,3}+0.01$} edge from parent node [left] {yes} }
        child { node [leaf, below right=1cm and 0cm of \tikzchildnode] {$Y_i = 0.98X_{i,4} - 0.03X_{i,3} - 0.01$} edge from parent node [right] {no} }
        edge from parent node [right] {no}
    };
\end{tikzpicture}
    \caption{Decision tree constructed by SMART. Numbers rounded to two decimal places. The RMSE on the true dataset with no noise is 0.065.}
    \label{fig:decision_tree_guess}
\end{figure}
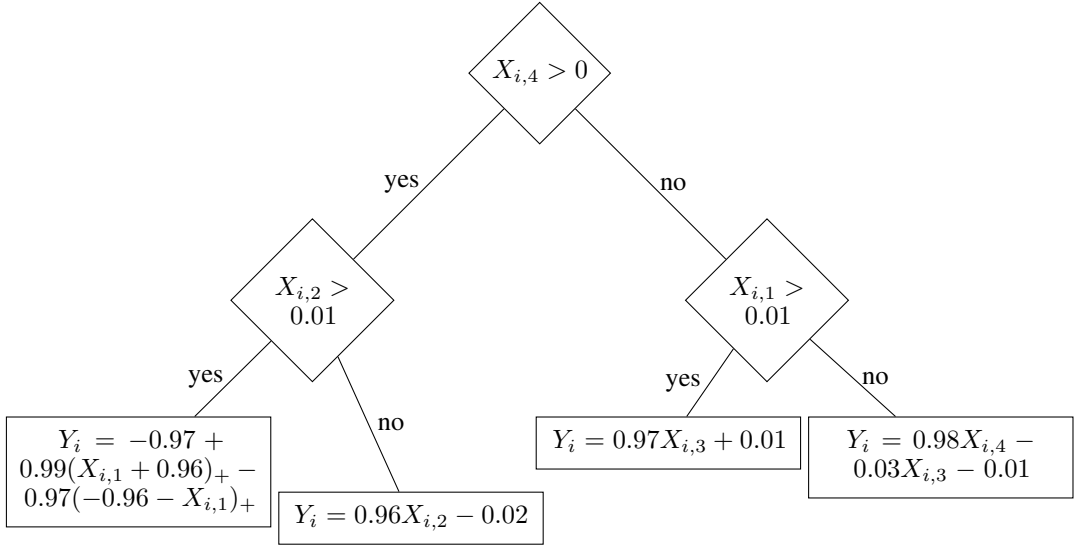

\par Figure \ref{fig:decision_tree_guess} shows the resulting fit of SMART on the dataset described above with a max degree of interaction of 1. The RMSE on the true function without noise is 0.065. The model was able to find all of the decision tree splits on the simulated data. Similarly, the model was able to adapt its structure through pruning to emphasize the most important variables at each leaf node. For instance, in the leftmost leaf node, the hinge equation simplifies to $0.99X_{i,1} - 0.02$ if $X_{i,1} > 0.96$ and $0.97X_{i,1} - 0.04$ if $X_{i,1} < 0.96$, both of which are very close to the true relationship of $X_{i,1}$. The rightmost leaf node was the only one to include an erroneous term, but the term had a very small coefficient of $-0.03$. This example shows that combining recursive tree splitting and pruning creates leaf nodes that can adapt to different relationships.

\section{Conclusion}
\label{sec:conclusion}

\par In this paper, we introduced SMART, a novel approach that integrates the strengths of Multivariate Adaptive Regression Splines with decision trees to model complex datasets characterized by continuous non-linearities and jump discontinuities. Through experiments on various simulated datasets, SMART demonstrated its effectiveness compared to state-of-the-art methods. Specifically, SMART's ability to combine tree-based splits with continuous spline functions allowed it to uniquely capture piecewise changes in functional form (Section \ref{subsec:PiecewiseTest}), while its native pruning mechanism enabled it to adapt to abrupt shifts in data patterns (Section \ref{subsec:PruningTest}). To encourage further adoption and development of SMART, we have provided an open-source implementation that is available for use and further refinement by researchers and practitioners \footnote{https://github.com/fyre87/SMART}. 

\section{Discussion}
\label{sec:discussion}
\par There is still room for improvement and future research on this topic. Researchers may be interested in incorporating ensemble techniques such as bagging with a tree of similarly complex nonlinear leaf nodes, potentially enhancing the robustness and accuracy of the method. Others may be interested in exploring multi-variable tree splits with this method. We also believe there are opportunities to develop more efficient approximation techniques for identifying the optimal tree split, which could lower the time complexity without compromising accuracy. Potential improvements such as parallelization of the QR algorithm or alternatives such as histogram-based splitting could be explored. These potential alternative techniques could enhance the accuracy and stability of the estimate over QR decomposition, which generally maintains numerical stability but may still accumulate roundoff errors when dealing with large datasets.

\newpage

\bibliographystyle{plainnat}
\bibliography{bibliography}

\appendix
\newpage
\section*{Appendix A: Hyperparameters}
\label{Appendix A}

\begin{table}[htbp]
\centering
\caption{Hyperparameters Considered in Grid Search}
\label{tab:grid_search_hyperparameters}
\begin{tabular}{|l|l|l|}
\hline
Model & Hyperparameter & Values Considered \\
\hline
Random Forest & max\_depth & 12, 15, 18, 21, 24, 27, 30 \\
 & max\_features & "sqrt", "log2", 1.0, 0.5 \\
\hline
XGBoost & n\_estimators & 100, 500, 1000 \\
 & max\_depth & 4, 6, 8 \\
 & learning\_rate & 0.01, 0.05, 0.1, 0.3 \\
 & subsample & 0.6, 0.8, 1.0 \\
 & reg\_lambda & 0.1, 1, 10 \\
\hline
Local Linear Forests & honesty & TRUE, FALSE \\
 & min\_node\_size & 3, 5, 7 \\
 & sample\_fraction & 0.4, 0.5, 0.6 \\
 & ll\_split\_lambda & 0.01, 0.1, 1 \\
\hline
\end{tabular}
\end{table}

\section*{Appendix B: Synthetic Equations}
\label{Appendix B}
Let
\[
P = \sum_{i=1}^{N} f_i(X)
\]
where there are three to five $f_i(X)$ of the form:
\[
f_i(X) = 
\begin{cases} 
a_iX_{*,i} \text{ } X_{*,i} \in \mathcal{U}(-10, 10)\text{ with probability $\frac{1}{3}$}  \\
a_iX_{*,i}^2 \text{ } X_{*,i} \in \mathcal{U}(-3, 3)\text{ with probability $\frac{1}{3}$}  \\
a_iX_{*,i}^3 \text{ } X_{*,i} \in \mathcal{U}(-2, 2)\text{ with probability $\frac{1}{3}$}  \\
\end{cases}
\]
one to two hinge terms of the form:
\[
f_i(X) = a_i(X_{*,i}-b_i)_+ - a^*_i(b_i - X_{*,i})_+
\]
with $b_i\in\mathcal{U}(-8, 8)$ and $X_{*,i}\in\mathcal{U}(-10, 10)$. One log term of the form:
\[
f_i(X) = a_i\text{log}(X_{*,i} + 1)
\]
and one interaction term of the form:
\[
f_i(X) = \frac{a_i}{4}X_{*,1}X_{*,2}
\]
where each coefficient $a_i \in \mathcal{U}(-4, -0.4) \cup \mathcal{U}(0.4, 4)$. Finally, we added a piecewise component by taking 10\% of the dataset by a random variable and shuffling all of the coefficients. This makes the final response of the equation:
\[
Y = c + \mathcal{N}(0, 10^2) + 
\begin{cases} 
P \text{ if } X_{*,j} \geq k \\
P^* \text{ if } X_{*,j} < k
\end{cases}
\]
where $P^*$ has the same structure as $P$ with all of the coefficients $a_i$ re-randomized. The intercept term is distributed as $c\in \mathcal{U}(-20, 20)$.
The five equations generated are determined by:
\[
Y_i = -1.2 + \epsilon + 
\begin{cases}
\begin{aligned}
    &-3.1X_{i,1}^3 + 2.1X_{i,2}^2 - 3.7X_{i,3}^2 + 2.0(X_{i,4}-1.2)_+ \\
    &+ 1.5(1.2-X_{i,4})_+ + 3.0\text{log}(X_{i,5} + 1) - 0.5X_{i,1}X_{i,2} 
\end{aligned} & \text{ if } X_{i,1} \leq 1.6 \\ 
\begin{aligned}
    &-3.9X_{i,1}^3 - 0.6X_{i,2}^2 + 2.9X_{i,3}^2 + 3.0(X_{i,4}-1.2)_+ \\
    &+ 1.3(1.2-X_{i,4})_+ + 2.4\text{log}(X_{i,5} + 1) + 0.3X_{i,1}X_{i,2} 
\end{aligned} & \text{ if } X_{i,1} > 1.6
\end{cases}
\]
\[
Y_i = 2.1 + \epsilon + 
\begin{cases}
\begin{aligned}
    &-2.7X_{i,1} + 1.3X_{i,2} - 1.9X_{i,3} + 2.7(X_{i,4}-2.4)_+ \\
    &- 2.4(1.2-X_{i,4})_+ + 2.2\text{log}(X_{i,5} + 1) - 0.2X_{i,1}X_{i,2} 
\end{aligned} & \text{ if } X_{i,2} \leq 8.0 \\ 
\begin{aligned}
    &3.7X_{i,1} + 3.6X_{i,2} - 2.0X_{i,3} - 3.2(X_{i,4}-2.4)_+ \\
    &+ 2.8(2.4-X_{i,4})_+ - 2.2\text{log}(X_{i,5} + 1) + X_{i,1}X_{i,2} 
\end{aligned} & \text{ if } X_{i,2} > 8.0
\end{cases}
\]
\[
Y_i = -4.3 + \epsilon + 
\begin{cases}
\begin{aligned}
    &-2.7X_{i,1}^3 + 2.7X_{i,2}^3 - 0.7X_{i,3} + 2.4(X_{i,4}-4.4)_+ + 0.4(4.4-X_{i,4})_+ \\
    &+ 2.3(X_{i,5}-2.9)_+ - 1.8(2.9-X_{i,5})_+ - 3.2\text{log}(X_{i,6} + 1) - 0.2X_{i,1}X_{i,2} 
\end{aligned} & \text{ if } X_{i,3} \leq 8.0 \\ 
\begin{aligned}
    &-1.6X_{i,1}^3 - 3.4X_{i,2}^3 - 2.9X_{i,3} - 1.0(X_{i,4}-4.4)_+ - 2.1(4.4-X_{i,4})_+ \\
    &+ 3.0(X_{i,5}-2.9)_+ - 1.8(2.9-X_{i,5})_+ + 3.9\text{log}(X_{i,6} + 1) - 0.2X_{i,1}X_{i,2} 
\end{aligned} & \text{ if } X_{i,3} > 8.0 \\ 
\end{cases}
\]
\[
Y_i = 6.7 + \epsilon + 
\begin{cases}
\begin{aligned}
    &-1.4X_{i,1}^2 - 3.7X_{i,2}^2 + 2.4X_{i,3} -3.2(X_{i,4}+2.8)_+ - 1.4(-2.8 -X_{i,4})_+ \\
    &- (X_{i,5}+4.4)_+ - 1.2(-4.4-X_{i,5})_+ - \text{log}(X_{i,6} + 1) + 0.8X_{i,1}X_{i,2} 
\end{aligned} & \text{ if } X_{i,1} \geq -2.4 \\ 
\begin{aligned}
    &1.4X_{i,1}^2 + 0.7X_{i,2}^2 + 2.8X_{i,3} - 3.4(X_{i,4}+2.8)_+ - 1.4(-2.8 -X_{i,4})_+ + \\
    &0.4(X_{i,5}+4.4)_+ - 2.5(-4.4-X_{i,5})_+ - 3.5\text{log}(X_{i,6} + 1) + 0.8X_{i,1}X_{i,2} 
\end{aligned} & \text{ if } X_{i,1} < -2.4 \\
\end{cases}
\]
\[
Y_i = 3.1 + \epsilon + 
\begin{cases}
\begin{aligned}
    &3.3X_{i,1}^2 - X_{i,2}^3 - 0.8X_{i,3} - 2.1X_{i,4} + 2.4X_{i,5} - 1.4(X_{i,6}+7.2)_+ \\
    &- 1.4(-7.2-X_{i,6})_+ - \text{log}(X_{i,7} + 1) - 0.9X_{i,1}X_{i,2} 
\end{aligned} & \text{ if } X_{i,5} \leq 8.0 \\ 
\begin{aligned}
    &-3.5X_{i,1}^2 - 3.2X_{i,2}^3 - 0.5X_{i,3} - 0.9X_{i,4} + 1.5X_{i,5} - 2.1(X_{i,6}+7.2)_+ \\
    &+ 3.3(-7.2-X_{i,6})_+ - 0.5\text{log}(X_{i,7} + 1) + 0.4X_{i,1}X_{i,2} 
\end{aligned} & \text{ if } X_{i,5} > 8.0 \\ 
\end{cases}
\]

\end{document}